\documentclass[sigconf,authorversion,nonacm]{acmart}

\AtBeginDocument{%
  \providecommand\BibTeX{{%
    \normalfont B\kern-0.5em{\scshape i\kern-0.25em b}\kern-0.8em\TeX}}}

\usepackage{graphicx}
\usepackage[utf8]{inputenc}
\usepackage{float}
\usepackage{cuted}

\usepackage{booktabs}
\usepackage{multirow}
\usepackage{graphicx}
\usepackage{tabularx}
\usepackage{wrapfig}
\usepackage{caption}
\usepackage{pgfplots}
\usepackage{pgfplotstable}

\usepackage{indentfirst}

\usepackage{microtype}

\begin{document}

\newcommand{\DeloitteNL}{Deloitte NL}
\newcommand{\DocQMiner}{DocQMiner}

\title{Green AI in Action: Strategic Model Selection for Ensembles in Production}

\author{Nienke Nijkamp}
\affiliation{%
    \institution{Delft University of Technology, Deloitte Risk Advisory BV}
    \country{The Netherlands}
}
\email{n.nijkamp@student.tudelft.nl}
\orcid{0009-0009-0627-1518}

\author{June Sallou}
\affiliation{%
    \institution{Delft University of Technology}
    \country{The Netherlands}
}
\email{J.Sallou@tudelft.nl}
\orcid{0000-0003-2230-9351}

\author{Niels van der Heijden}
\affiliation{%
    \institution{University of Amsterdam, Deloitte Risk Advisory BV}
    \country{The Netherlands}
}
\email{n.vanderheijden@uva.nl}
\orcid{0009-0005-0632-8327}

\author{Luís Cruz}
\affiliation{%
    \institution{Delft University of Technology}
    \country{The Netherlands}
}
\email{L.Cruz@tudelft.nl}
\orcid{0000-0002-1615-355X}

\date{\today}

\begin{abstract}

Integrating Artificial Intelligence (AI) into software systems has significantly enhanced their capabilities while escalating energy demands. Ensemble learning, combining predictions from multiple models to form a single prediction, intensifies this problem due to cumulative energy consumption. 

This paper presents a novel approach to model selection that addresses the challenge of balancing the accuracy of AI models with their energy consumption in a live AI ensemble system. We explore how reducing the number of models or improving the efficiency of model usage within an ensemble during inference can reduce energy demands without substantially sacrificing accuracy.

This study introduces and evaluates two model selection strategies, \textit{Static} and \textit{Dynamic}, for optimizing ensemble learning systems' performance while minimizing energy usage.  Our results demonstrate that the \textit{Static} strategy improves the F1 score beyond the baseline, reducing average energy usage from 100\% from the full ensemble to 62\%. 
The \textit{Dynamic} strategy further enhances F1 scores, using on average 76\% compared to 100\% of the full ensemble. 

Moreover, we propose an approach that balances accuracy with resource consumption, significantly reducing energy usage without substantially impacting accuracy. This method decreased the average energy usage of the \textit{Static} strategy from approximately 62\% to 14\%, and for the \textit{Dynamic} strategy, from around 76\% to 57\%.

Our field study of Green AI using an operational AI system developed by a large professional services provider shows the practical applicability of adopting energy-conscious model selection strategies in live production environments.
\end{abstract}

\keywords{Green AI, Ensemble Learning, Model Selection, Information Extraction}

\ccsdesc[500]{Software and its engineering~Software creation and management}
\ccsdesc[500]{Computing methodologies~Artificial intelligence}
\ccsdesc[100]{Applied computing~Document management and text processing}

\maketitle
\vspace{-0.5em}
\section{Introduction}
Over recent years, the integration of Artificial Intelligence (AI) into modern software systems, commonly known as AIware or AI-powered software~\cite{monett2021ai}, has experienced significant growth~\cite{georgiou2022green}. As a result, the demand for resources, particularly the energy needed for training, deploying, and running inferences with these AI models, has surged considerably~\cite{strubell2019energy, anthony2020carbontracker, chien2023reducing}. To illustrate the resource consumption associated with inference, a ChatGPT-like application handling 11 million requests per hour is estimated to emit 12,800 tons of CO$_2$ annually, making inference 25 times more carbon-intensive than training GPT-3~\cite{chien2023reducing}. 

Ensemble learning, a method that combines multiple models to create a more effective solution, while highly effective, tends to intensify this energy consumption problem due to the cumulative energy requirements of the individual models~\cite{li2023towards, david2021adaptive}. 

The focus of the AI research community has predominantly been on improving the accuracy of AI models, overlooking the significant energy costs associated with them. AI's rising environmental and financial cost has led to a pressing need for a more balanced approach to AI development that considers accuracy and energy efficiency~\cite{strubell2019energy}. The emerging field of Green AI addresses this gap, promoting a favourable trade-off between efficiency and accuracy~\cite{schwartz2020green}. A significant gap remains in implementing Green AI principles within the industry~\cite{verdecchia2023systematic}. To bridge this gap, we have partnered with \DeloitteNL~\cite{DeloitteNetherlands2024}, a large professional services network specializing in, amongst others, digital risk solutions, to carry out a field experiment on an active AI system.

Our research explores the intersection of Green AI and ensemble learning in a production environment, a domain that has yet to be thoroughly investigated. Simply running all models every time is not an efficient strategy~\cite{zhou2002ensembling}. The challenge is finding a more intelligent approach for running inference with ensemble learning~\cite{whitaker2022prune}, reducing energy consumption while maintaining or improving accuracy.

In this paper, we propose a solution that involves a selective approach to using models within an ensemble. The core of this approach is the concept of \textit{model selection strategies}, which refers to various methods of selecting specific subsets of models for individual tasks rather than using the complete set of models for every task~\cite{caruana2004ensemble}. Our goal in implementing these model selection strategies is to balance achieving accuracy and managing computational costs. This goal leads to the following research question. \newline
\vspace{-0.5em}

\textit{Research Question:} What are the impacts of implementing model selection strategies on the accuracy and energy usage of ensemble learning systems? \newline

\vspace{-0.5em}

Two model selection strategies, \textit{Static} and \textit{Dynamic}, are investigated for optimizing model performance. Both strategies start by evaluating all subsets of the entire ensemble on a validation set, selecting the combination with the highest accuracy. \textit{Static} selection identifies the best overall model selection, while \textit{Dynamic} selection chooses the best selection per specific property within the domain. 

Additionally, we consider the computational cost per model by employing a metric that discounts accuracy with energy consumption. This method is incorporated into both the \textit{Static} and \textit{Dynamic} selection strategies, ensuring a balanced consideration of performance and resource efficiency.

This empirical study explores the potential of model selection strategies within an ensemble of AI models. We conduct a field study on \DocQMiner~\cite{DeloitteDocQMiner}, a live, industry-used AI system for text information extraction from large volumes of documents, to assess the practical implications of our proposed model selection strategies. 



This research provides practical insights into making model ensembles more efficient by 
examining and implementing our model selection strategies within an ensemble learning context. 
Our contributions to the field of AI in software systems using ensemble learning are the following: 
\begin{enumerate}
    \item A detailed evaluation of \textit{Static} and \textit{Dynamic} model selection strategies in a production environment.
    \item An approach to enhance these strategies by incorporating energy usage metrics, significantly lowering energy consumption. 
\end{enumerate}
 
These contributions demonstrate that model selection strategies not only significantly reduce resource consumption but also have the potential to maintain or increase the accuracy of the AI system.

Moreover, these findings highlight the practicality and necessity of integrating Green AI principles into AI development, working towards more sustainable and efficient AI applications in the industry\footnote{The replication package for this study is available at the following DOI link: \url{https://doi.org/10.6084/m9.figshare.25481269}}.


\vspace{-0.5em}
\section{Background}

In this section, we highlight the concept of ensemble learning, followed by a detailed study of the use case and the infrastructure of the AI system. 

\subsection{Ensemble Learning}

Ensemble learning is a strategic approach that combines several individual and diverse models to achieve better generalization performance~\cite{zhou2012ensemble}. The strength of ensemble learning lies in its diversity; a combination of models can provide a more robust and accurate output than any single model can~\cite{david2021adaptive, ganaie2022ensemble}. Popular applications of ensemble learning are speech recognition~\cite{deng2012use, li2017semi} and classification~\cite{tur2012towards, palangi2014recurrent}, image classification~\cite{wang2020particle, li2019semi, huang2017snapshot} and forecasting~\cite{qiu2014ensemble, liu2017flood, carta2021multi, singla2022ensemble}. 

A well-known method within ensemble learning is bagging (Bootstrap Aggregating), which enhances the stability and accuracy of machine learning algorithms by training multiple learners on various subsets of the original dataset and then aggregating their predictions to form a final decision~\cite{breiman1996bagging}.


At the core of our research are \textit{model selection strategies} for ensembles. Model selection entails selecting a subset of models from an ensemble of models to optimize for a particular performance metric~\cite{caruana2004ensemble}. We use \textit{model selection} within ensembles to reduce resource consumption in alignment with Green AI principles. 

\subsection{Use case}

To evaluate the impact of using \textit{model selection strategies} for ensemble learning in a live production environment, we study the live AI system \DocQMiner \space~\cite{DeloitteDocQMiner}, a proprietary tool owned and developed by \DeloitteNL. 
This system uses diverse machine learning (ML) models and NLP technologies to extract, process, and analyze data from textual documents. It is an information extraction tool widely used in the industry across many domains, with document set sizes ranging from 100 to over 100,000 documents. 
This paper focuses on the use case of extracting relevant properties from contracts.  

\begin{figure}[htbp]
    \centering
    \includegraphics[width=0.80\columnwidth]{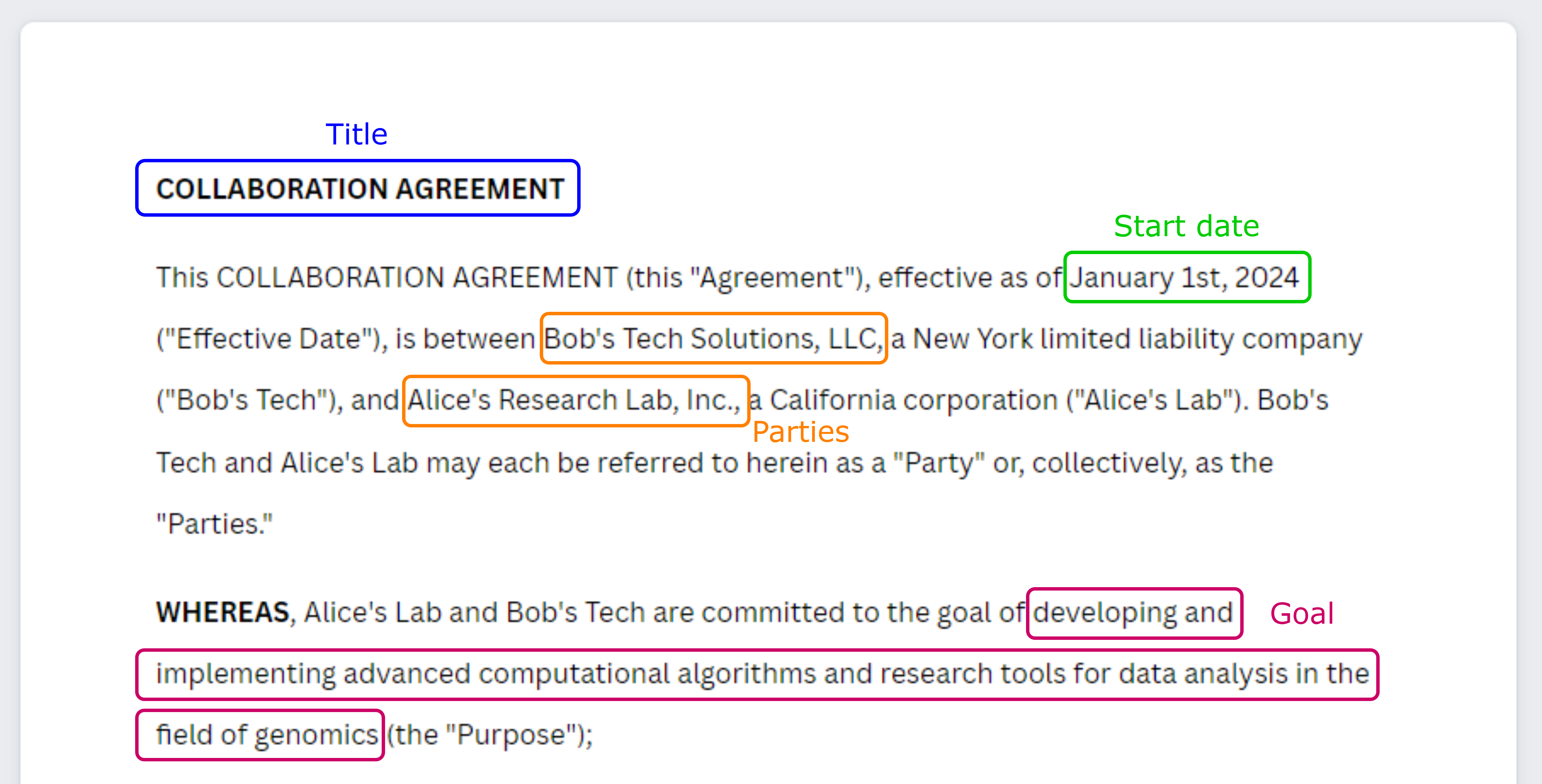}
    \caption{Example of a document with relevant properties highlighted}
    \label{fig:example documnet}
\end{figure}
After processing a document, the AI system makes predictions for predefined key properties. These properties are chosen by the user according to their use case, as shown in an example in Figure \ref{fig:example documnet} where properties such as \textit{Title}, \textit{Parties}, and \textit{Goal} in a contract are highlighted. The tool allows users to input a contract and delivers the predictions for the properties of interest. This design makes contract review more efficient, especially for complex documents~\cite{nawar2022open}.

\begin{figure}[htbp]
    \centering
    \includegraphics[width=0.75\columnwidth]{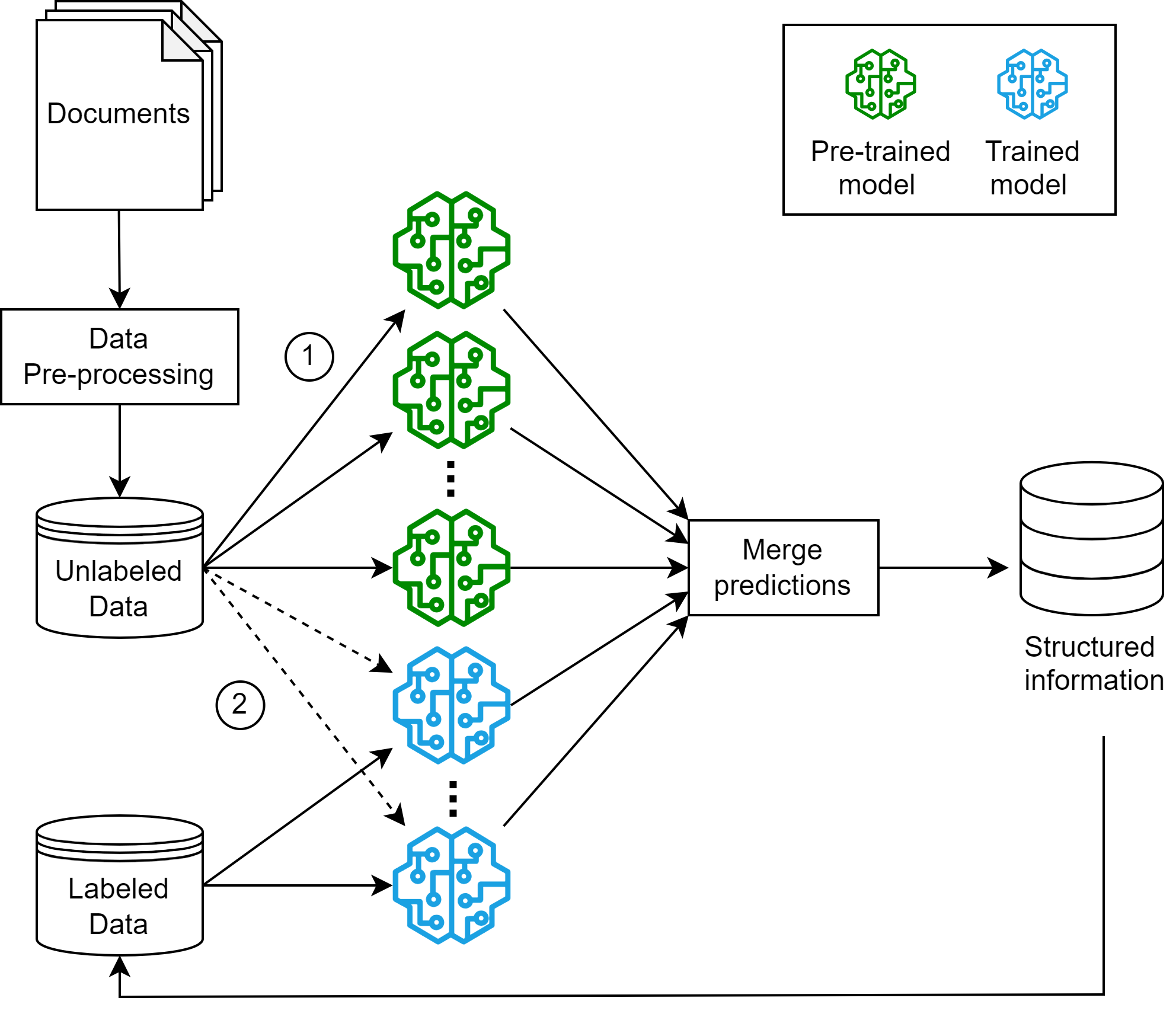}
    \caption{Workflow of \DocQMiner. (1) Initially, documents are processed using the pre-trained models. After processing an initial set of documents, models are trained using the processed data. (2) After training models, documents are processed using the pre-trained and trained models.}
    \label{fig:models}
\end{figure}

\DocQMiner \space employs diverse (pre-)trained models, as shown in Figure \ref{fig:models}, to compile a ranked top 5 of textual predictions for every queried property. The AI system deploys an ensemble framework that utilizes the bagging approach. Multiple models that are independently trained on randomly generated subsets of data are combined to produce predictions. 

For every property, each model in the ensemble produces a set of predictions aggregated to form the set of the five final predictions. The user selects the correct prediction, or, in case the desired answer is not part of the prediction, the user manually selects the answer from the document. This human-in-the-loop workflow ensures the validity of the processed data. 

The system subjects input documents to a pre-processing step to ensure that the text within is prepared for subsequent analysis by its models.
Out of the box, the ensemble contains a set of pre-trained models to make predictions for the properties, indicated by the (1) in Figure \ref{fig:models}. After the system has processed a set of documents within a domain, it can train an additional set of models using the in-domain data from those documents. Subsequently, the system processes documents using both the pre-trained models and those trained on the processed documents, as depicted by (2) in Figure \ref{fig:models}.

\subsection{Resource consumption}

This paper analyses the energy consumption of the AI system in a production environment. Therefore, we leverage the existing monitoring tool within the system, Datadog~\cite{DataDog2024}, which offers an extensive set of features for extracting metrics from the system. This cloud monitoring service provides a framework for tracking and analyzing energy use metrics within our AI deployments, as noted in industry literature~\cite{mckaig2022metrics}.

\DocQMiner \space uses a substantial amount of CPU when processing a single document. Considering \DocQMiner \space processes document sets ranging in size from 100 to over 100,000 documents per instance, there are environmental and financial implications. This energy usage escalates operational costs and enlarges the carbon footprint of using \DocQMiner, challenging the sustainable principles of \DeloitteNL~\cite{Deloitte2023EnvironmentalImpacts}.

\section{Model selection}

This study aims to analyze the impact of implementing model selection strategies in a live AI system that uses an ensemble of models on its energy usage and accuracy. In this section, we highlight how two selection strategies, \textit{Static}~\cite{luo2016review} and \textit{Dynamic}~\cite{cordeiro2023post}, can be used for this, and we describe our approach to using them to reduce resource consumption even more efficiently, \textit{Energy-Aware selection}.







\subsection{Model selection for energy efficiency}
In pursuit of sustainable AI practices, our study assesses existing \textit{Static}~\cite{luo2016review} and \textit{Dynamic}~\cite{cordeiro2023post} model selection strategies to reduce energy consumption in a live AI system. These strategies are used to reduce the number of models or improve the efficiency of model usage used during inference, which is a significant determinant of overall energy usage~\cite{li2023towards}. 

The \textit{Static} strategy selects an optimal subset of models for general tasks across the domain, while the \textit{Dynamic} strategy adapts model selection to the specifics of each task, aiming to conserve energy without compromising the system's accuracy.

Our main contribution lies in the novel \textit{Energy-Aware selection} approach, which enhances the standard \textit{Static} and \textit{Dynamic} strategies by integrating an energy-aware metric in their application. This metric informs the selection process, ensuring that only the most energy-efficient models are chosen for the task at hand.

\subsection{Static selection} 
\label{subsubsec:static}
\textit{Static} selection involves choosing the optimal subset of models for an entire domain. This approach generalizes the unique characteristics of the domain and selects the optimal subset across all queried properties. The following equation shows the process of \textit{Static} selection.
\begin{equation}
    S_{a^*} = arg max_{S_i \in S} F1(S_i)
\end{equation}
where $S$ is the set of all possible model subsets, F1($S_i$) is the F1 score of subset $S_i$ on the training set, and $S_{a^*}$ as the optimal model subset chosen for evaluation.

\subsection{Dynamic selection} 
\label{subsubsec:dynamic}
\textit{Dynamic} selection is based on the belief that a model subset might not be optimal for an entire domain, but more specifically for the properties within the domain~\cite{cordeiro2023post}. Therefore, we take the optimal subset for every queried property within the domain. This approach could be beneficial as the selection is more optimized per property. The process of \textit{Dynamic} selection is represented by Equation~\ref{eq:dyn_selec}.
\begin{equation}\label{eq:dyn_selec}
    S_{a^*}(p_j) = arg max_{S_i \in S} F1(S_i, p_j)
\end{equation}
We note $P$ as the set of all properties within the domain. $F1(S_i, p_j)$ is the F1 score of subset $S_i$ for property $p_j$ on the training set. $S_{a^*}(p_j)$ is the optimal model subset for property $p_j$ in $P$ chosen for evaluation.

\subsection{Energy-Aware selection}
\label{subsubsec:includingcost}
Both selection strategies, \textit{Static} and \textit{Dynamic}, should reduce energy consumption because a subset of models is used instead of all. However, neither strategy considers how different models compare in terms of energy efficiency. For instance, one model could slightly improve accuracy while costing a significantly larger amount of energy than another. 

We propose an enhancement to the \textit{Static} and \textit{Dynamic} approach that discounts accuracy with resource consumption in the selection of models, \textit{Energy-Aware selection}. We use the GreenQuotientIndex (GQI)~\cite{gowda2023watt} to factor the trade-off between accuracy and electricity usage. We add GQI to both \textit{Static} and \textit{Dynamic} versions, and compute it as follows: 
\vspace{-3mm}
\begin{equation}
    GQI_{static} = \beta \times \frac{F1(S_i)^\alpha}{log_{10}(C(S_i))}
\end{equation}
\begin{equation}
    GQI_{dynamic} = \beta \times \frac{F1(S_i,p_j)^\alpha}{log_{10}(C(S_i))}
\end{equation}
This metric evaluates the trade-off between accuracy and power consumption, where $\alpha$ and $\beta$ are constants used to scale the GQI. The power consumption ($C(S_i)$) can vary significantly across different models, and therefore the logarithm of the power consumption is taken. Not all accuracy points ($F1(S_i)$ for $GQI_{static}$ and $F1(S_i, p_j)$ for $GQI_{dynamic}$) have the same weight, as it is much easier to get from 0.4 to 0.5 then it is to go from 0.8 to 0.9. Therefore, the power (constant $\alpha$) of the accuracy reflects the difference in difficulty.

Through the previously mentioned monitoring tool, Datadog~\cite{DataDog2024}, there is no availability for power consumption. We, therefore, use CPU usage as a proxy to discount the accuracy. 
We use this way of discounting the accuracy scores for both the \textit{Dynamic} and the \textit{Static} selection approaches highlighted above.

\section{Methodology}

This section outlines the methodology and evaluation of our \textit{model selection strategies}. 
We start with data collection and analysis, followed by the experimental setup. We continue with the performance evaluation metrics. Lastly, we explore the energy consumption of the models during inference. 
\subsection{Data}

The datasets for our experiments are selected based on a set of criteria. Each document within the dataset must contain contracts with a complete text, its associated properties, and annotated responses corresponding to these properties. We employ open-source datasets for our study, enhancing the reproducibility of our results. The CUAD~\cite{hendrycks2021cuad} dataset and a dataset from the work of Leivaditi et al.~\cite{leivaditi2020benchmark} are used in this study, both of which were designed to optimize contract review processes and improve the effectiveness of information extraction.

\subsubsection{CUAD}

The Contract Understanding Atticus Dataset~\cite{hendrycks2021cuad} (CUAD) is an extensive corpus tailored for commercial legal contract analysis. It comprises over 13,000 labels from 510 contracts divided into 25 contract types. Each document contains one or more of 41 distinct properties. The dataset presents a challenging research benchmark that can be used to enhance deep learning models' performance for contract analysis/understanding~\cite{chang2023survey}.

\subsubsection{Lease Contracts}

Leivaditi et al.~\cite{leivaditi2020benchmark} introduced a specialized benchmark dataset focused on lease agreement documents. These documents were sourced from a publicly available dataset by the U.S. Securities and Exchange Commission (SEC, 2020). The dataset concentrates on extracting specific properties, including information about the lessor and details of the leased space.

{
\noindent 
\captionof{table}{Relevant characteristics of CUAD and Lease Agreement dataset}\label{tab:analysis}
\small
\begin{tabularx}{\columnwidth}{@{}Xrr@{}}
\toprule
Dataset & CUAD~\cite{hendrycks2021cuad} & \begin{tabular}[c]{@{}l@{}}Lease \\ Agreement~\cite{leivaditi2020benchmark}\end{tabular} \\
\midrule
\#Documents & 510 & 123 \\
\#TypesOfDocuments & \textbf{25} & \textbf{1} \\
\#WordsPerDocument & 7861 & 8053 \\
\#AnnotatedProperties & 41 & 12 \\
\#TotalQueries & 20,910 & 1476 \\
\#MissingAnnotations & 13,959 & 494 \\
MissingAnnotation (\%) & \textbf{66.77} & \textbf{33.47} \\
\bottomrule
\end{tabularx}
}

\normalsize 
\subsubsection{Analysis of datasets} 
We compare and analyze the datasets employed to understand the characteristics of the datasets and ensure the validity of the results. Table \ref{tab:analysis} shows the comparison of relevant characteristics. The CUAD and Lease Agreement datasets exhibit similar document lengths, with an average of approximately 8,000 words per document, as indicated by the \textit{\#WordsPerDocument} metric. Consequently, we will not regard document length as influencing our results. 

The CUAD dataset encompasses a diverse collection of 25 types of contracts, reflecting a wide range of legal agreements. In contrast, the Lease Agreement dataset focuses solely on a single type of contract, specifically lease agreements. With the CUAD dataset's variety, the model selection process must account for the nuances and intricacies inherent in different types of contracts. Conversely, the Lease Agreement dataset's singular focus may allow for more specialized model tuning and optimization tailored specifically to lease agreements. 

Additionally, there is a notable difference in the number of annotated properties between the two datasets. With \textit{\#MissingAnnotations}, we indicate the number of properties that were queried but did not have a ground truth in the document, thus missing an annotation. Specifically, 66.77\% of the queried properties in the CUAD dataset lack annotations, in contrast to the Lease Agreement dataset, which is missing annotations for only 33.47\% of its properties. Despite the significant number of missing annotations, we use these datasets to cover a broader array of use cases, including documents where the answer might not always exist. The varied percentage of missing annotations allows us to cover more ground in our results.
\vspace{-0.3em}
\subsection{Experimental setup}

To integrate models tailored for specific domains, we select a subset of documents from the CUAD~\cite{hendrycks2021cuad} and Lease Agreement~\cite{leivaditi2020benchmark} datasets that reflect the entire dataset. We process, annotate, and train models on these documents. After training, we run an evaluation on a held-out test set.

For both approaches, we run inference using the complete model ensemble on the documents from the validation set. We then evaluate the performance of the entire ensemble and each subset of models within the ensemble. The subset that demonstrates the highest performance, as determined by F1 scores from the training set, is selected for further testing.

This optimally performing subset is applied to the test set documents to evaluate its effectiveness on new data. To ensure the validity of the results, we create the validation and test datasets through 5-fold cross-validation.

As implemented in the tool, the predictions made by the entire model ensemble establish the performance evaluation baseline.
\vspace{-0.3em}
\subsection{Performance Evaluation}

When paired with the annotations, the ensemble's predictions can result in various scenarios: True Positive, where the prediction aligns with the annotation; False Positive, where a prediction exists but fails to match the annotation; and False Negative, where a prediction exists but no corresponding annotation for the property exists. 

To evaluate the performance of the strategies and compare it to the baseline situation of using all the models, we use the F1 score at k. F1 score is a balanced metric of Recall and Precision~\cite{powers2020evaluation}.

\emph{Precision}: Precision~\cite{sokolova2009systematic} is the percentage of all predictions that match an annotation for the top k predictions: 
\small
\begin{equation}
    Precision@k = \frac{TP}{TP + FP} \text{ \space \space for the top k predictions}
\end{equation}
\normalsize
\emph{Recall}: Recall~\cite{sokolova2009systematic} is the percentage of correctly predicted annotations for the top k predictions: 
\small
\begin{equation}
    Recall@k = \frac{TP}{TP + FN} \text{ \space \space for the top k predictions}
\end{equation}
\normalsize
Precision discounts for the number of False Positives and Recall for the number of False Negatives. Increasing Precision typically reduces Recall, and vice versa: increasing Recall decreases Precision. This trade-off shows the importance of prioritizing one over the other based on specific objectives. 

\emph{F1 score}: F1 score provides a balanced measure of Precision and Recall for the top k predictions. To ensure the robustness and reliability of our research, we select a subset of models based on the F1 score. 
\small
\begin{equation}
    F1@k = 2 \cdot \frac{Precision@k \cdot Recall@k}{Precision@k + Recall@k}
\end{equation}
\normalsize
\DocQMiner \space~\cite{DeloitteDocQMiner} only shows the top 5 predictions produced by the set of models. Consequently, we evaluate the results based on k set to five. F1@5 serves as our primary criterion for selecting among different strategies, while we report Precision@5 and Recall@5  to provide a detailed view of the factors contributing to the F1@5 score.

\subsection{Resource consumption during inference}

To find the most energy-efficient model subset, we need a measurement of the models' consumption during inference. 
\DocQMiner \space is a live-production environment; therefore, we want to measure the system's resource consumption live. Due to this limitation, we focused on tracking the Central Processing Unit (CPU) usage. With the use of Datadog~\cite{DataDog2024}, we can record the CPU usage per second for each process, which allows us to isolate the CPU usage per second, specifically during the inference phase of the models. By taking the cumulative sum of the CPU usage per second over the total duration of the process, we obtain the CPU seconds per process~\cite{brady2005virtualization}: 
\begin{equation}
    \textbf{CPU seconds} = \sum_{i=1}^{n} u_i 
\end{equation}
$n$ is the total number of seconds for the process, and $u_i$ is the CPU usage per second during $i$. Datadog~\cite{DataDog2024} performs this sum calculation. 

This setup shows a clear view of each model's performance, given that any other model does not influence the CPU utilization of one model. We gather the data for processing each document and each model for a set of intervals of the amount of queried properties.

Given that these measurements are taken in a live production environment, we designed our approach to yield results that closely represent the most probable outcomes. We acknowledge the inherent variability of a production environment, so we plot the measurements' outcomes to account for the variance in results. To ensure the validity of the collected data, we repeat the measurement per number of queried properties 30 times. 

\vspace{-0.5em}
\section{Results}

\begin{table*}[]
\caption{Results of \textit{Full Ensemble}, \textit{Static} and \textit{Dynamic} strategy in Precision@5, Recall@5, F1@5, Number of Correct Prediction, Number of Incorrect predictions and Consumption in percentage for CUAD and Lease Agreement dataset}
\resizebox{\textwidth}{!}{%
\begin{tabular}{@{}llcccccccccccc@{}}
\toprule
 &  & \multicolumn{6}{c}{CUAD~\cite{hendrycks2021cuad}} & \multicolumn{6}{c}{Lease Agreement~\cite{leivaditi2020benchmark}} \\ \cmidrule(r){3-8} \cmidrule(l){9-14}
\multicolumn{2}{l}{Strategy} & P@5 & R@5 & F1@5 & \begin{tabular}[c]{@{}c@{}}\# Correct \\ predictions\end{tabular} & \begin{tabular}[c]{@{}c@{}}\# Incorrect\\ predictions\end{tabular} & \begin{tabular}[c]{@{}c@{}}Relative \\ CPU Usage\end{tabular} & P@5 & R@5 & F1@5 & \begin{tabular}[c]{@{}c@{}}\# Correct \\ predictions\end{tabular} & \begin{tabular}[c]{@{}c@{}}\# Incorrect\\ predictions\end{tabular} & \begin{tabular}[c]{@{}c@{}}Relative \\ CPU Usage\end{tabular} \\ \cmidrule(r){1-8} \cmidrule(l){9-14}
\multicolumn{2}{l}{\textit{Full Ensemble}} & \textit{0.0871} & \textit{0.5338} & \textit{0.1495} & \textit{490} & \textit{5928} & \textit{100\%} & \textit{0.1386} & \textit{0.4448} & \textit{0.2203} & \textit{90} & \textit{584} & \textit{100\%} \\ \midrule
\multirow{2}{*}{Static} & Original & 0.6457 & 0.4872 & 0.5544 & 408 & 265 & 60.39\% & 0.1763 & 0.3677 & 0.2545 & 77 & 376 & 64.62\% \\
 & Energy-Aware & 0.6370 & 0.4652 & 0.5366 & 394 & 259 & 26.52\% & 0.1817 & 0.2404 & 0.2054 & 57 & 278 & 0.82\% \\ \midrule
\multirow{2}{*}{Dynamic} & Original & 0.6118 & 0.5448 & 0.5750 & 446 & 322 & 71.28\% & 0.1852 & 0.4732 & 0.2652 & 93 & 427 & 81.22\% \\
 & Energy-Aware & 0.6099 & 0.5471 & 0.5756 & 446 & 326 & 67.11\% & 0.2106 & 0.3415 & 0.2588 & 70 & 282 & 47.43\% \\ \bottomrule
\end{tabular}%
}
\end{table*}

Our evaluation of selection strategies across two datasets—CUAD and Lease Agreement—reveals significant differences in accuracy metrics, including Precision at 5 (P@5), Recall at 5 (R@5), F1 score, and the number of correct and incorrect predictions made by the models involved (see Table 2). \textit{Full Ensemble} reflects the baseline of using all the models for every property within all documents.
\vspace{-0.2em}
\subsection{Experimental context}
\paragraph{Legal information extraction} Extracting relevant information from legal documents presents significant challenges due to their complex nature. These texts often require identifying specific details within extensive documents. This task notably differs from more straightforward tasks like classification, where an F1 score below 0.6 might be deemed insignificant. In the context of legal text analysis, the F1 scores are typically lower, reflecting the intricate nature of the work involved.

These lower scores are supported in research conducted by Savelka et al.~\cite{savelka2023can} using GPT-4 for property extraction from complex legal documents, which reported a Precision of 0.63, a Recall of 0.46, and an F1 score of 0.53. Despite being from a different dataset, these results show that relevant information extraction from legal documents is not a trivial task, and the results from the \textit{Full Ensemble} should be interpreted as such. 
\vspace{-0.5em}
\paragraph{Recall-oriented system} \DocQMiner \space~\cite{DeloitteDocQMiner} is developed prioritizing Recall as the key metric, which aligns with its human-in-the-loop design. This design allows users to choose the best answer from the predictions provided. Consequently, the model development concentrates on accurately identifying relevant properties rather than minimizing the prediction of incorrect properties. 
\vspace{-0.5em}
\subsection{Full Ensemble}
For CUAD, the \textit{Full Ensemble} strategy achieved a P@5 of 0.0871, R@5 of 0.5338, and F1 of 0.1495, making 490 correct and 5928 incorrect predictions, with 100\% CPU usage.
For Lease Agreement, \textit{Full Ensemble} had a P@5 of 0.1386, R@5 of 0.4448, and F1 of 0.2203, with 90 correct and 584 incorrect predictions, at 100\% CPU usage. 
\vspace{-0.3em}
\subsection{Static Strategy}
For CUAD, the \textit{Static Original} variant, as defined in Section \ref{subsubsec:static}, achieved a P@5 of 0.6457, R@5 of 0.4872, and F1 of 0.5544, with 408 correct and 265 incorrect predictions, consuming 60.39\% CPU compared to the \textit{Full Ensemble}.
For Lease Agreement, \textit{Static Original} posted a P@5 of 0.1763, R@5 of 0.3677, and F1 of 0.2545, with 77 correct and 376 incorrect predictions, at 64.62\% CPU usage. 

The \textit{Static Energy-Aware} variation, as defined in Section \ref{subsubsec:includingcost}, for CUAD showed a slight decrease in performance to an F1 of 0.5366, P@5 of 0.6370, R@5 of 0.4652, with 394 correct and 259 incorrect predictions, and reduced energy consumption to 26.52\% CPU. For the Lease Agreement dataset, it managed an F1 of 0.2054, P@5 of 0.1817, and R@5 of 0.2404, with 57 correct and 278 incorrect predictions, drastically cutting CPU use to 0.82\%. 

Compared to the \textit{Full Ensemble} baseline, overall Recall has slightly declined; however, overall Precision has increased, especially for the CUAD dataset. These results show that the \textit{Static} strategy can correctly identify many properties while reducing the `noise' of incorrect predictions.

\subsection{Dynamic Strategy}
For CUAD, the \textit{Dynamic Original} strategy, as defined in Section \ref{subsubsec:dynamic}, resulted in an F1 score of 0.5750, a P@5 of 0.6118, an R@5 of 0.5448, 446 correct and 322 incorrect predictions, at 71.28\% CPU consumption.
For Lease Agreement, it achieved an F1 score of 0.2652, a P@5 of 0.1852, an R@5 of 0.4732, 93 correct and 427 incorrect predictions, with 81.22\% CPU usage compared to the \textit{Full Ensemble}.

The \textit{Dynamic Energy-Aware}, defined in Section \ref{subsubsec:includingcost}, showed for CUAD an F1 of 0.5756, P@5 of 0.6099, and R@5 of 0.5471, with 446 correct and 326 incorrect predictions, lowering CPU usage to 67.11\%. For the Lease Agreement dataset, the F1 was 0.2588, P@5 of 0.2106, R@5 of 0.3415, with 70 correct and 282 incorrect predictions, reducing CPU consumption to 47.43\%.

Overall, the \textit{Dynamic Original} strategy outperforms the baseline on both Precision and Recall. The increase in Recall is notable, considering \DocQMiner \space is tailored to Recall. The increase in Recall is suspected to be due to how the ensemble `dilutes' the predictions, and the \textit{Dynamic} strategy specializes in specific properties. 

\subsection{Strategy selection}

Identifying an optimal strategy for an ensemble of models hinges on a few considerations. The evaluation of the \textit{Full Ensemble}, \textit{Static}, and \textit{Dynamic} strategies across the CUAD and Lease Agreement datasets provides valuable insights into their respective strengths and weaknesses. 
\vspace{2em}
\subsubsection{Precision - Recall trade-off} 
\paragraph{For Precision} The \textit{Static Original} strategy stands out in the CUAD dataset with a Precision (P@5) of 0.6457 and an F1 score of 0.5544, significantly reducing the noise of incorrect predictions. Similarly, the Lease Agreement dataset performs with a Precision of 0.1763, compared to 0.1386 Precision for the \textit{Full Ensemble}. These numbers suggest that the \textit{Static Original} strategy offers a solution when minimising false positives is the goal.

\paragraph{For Recall} The \textit{Dynamic Original} strategy shines by delivering a Recall (R@5) of 0.5448 for CUAD and 0.4732 for Lease Agreement, coupled with the highest F1 scores (0.5750 and 0.2652). This strategy ensures that more relevant properties are captured, making it ideal when maximising the number of accurately identified properties, which is the goal.

\subsubsection{Resource efficiency} 
In a resource constraint environment, their performance and efficiency balance should inform the choice between the \textit{Static Energy-Aware} and \textit{Dynamic Energy-Aware} strategies.
\vspace{-0.5em}
\paragraph{Extreme Efficiency} The \textit{Static Energy-Aware} strategy is unparalleled, especially evident in the Lease Agreement dataset, with CPU usage reduced to nearly 1\%. This strategy is suitable for projects where every bit of computational resource saved makes an impact, even at the expense of some accuracy.
\vspace{-0.5em}
\paragraph{Balanced Approach} The \textit{Dynamic Energy-Aware} strategy, while not as resource-efficient as its \textit{Static} counterpart, offers a balance between accuracy and resource usage. This balance makes it ideal for scenarios where a moderate resource reduction is acceptable if it means retaining a higher level of accuracy.

\subsubsection{Specificity of properties}
If the task involves identifying particular properties within legal documents, the specialisation afforded by the \textit{Dynamic} strategies might yield better results. The \textit{Dynamic} strategies are fine-tuned to identify specific properties more effectively, possibly at the cost of broader applicability that \textit{Static} strategies can offer. 

However, the success of the \textit{Dynamic} approach hinges on the availability of sufficient information about the specific properties within the test set to identify the optimal subset accurately. In cases where such specific information is unavailable, the \textit{Static} strategy may be the more suitable option, balancing the need for broader coverage with the available data.

\subsection{Impact of selection strategies}

Our results indicate that implementing model selection strategies can significantly impact both the accuracy and resource consumption of ensemble learning systems. The evaluation of the \textit{Static} strategy suggests a notable improvement in Precision, reducing the number of incorrect predictions. In comparison, the \textit{Dynamic} strategy excels in Recall, effectively retrieving more relevant instances while achieving a reduction in energy consumption, making it ideal for applications where capturing as much relevant information as possible is critical.

Furthermore, by enhancing these strategies with the cost-inclusive GreenQuotientIndex (GQI)~\cite{gowda2023watt}, we have demonstrated a method for reducing the models' energy consumption without substantially sacrificing accuracy. For instance, the \textit{Static Energy-Aware} strategy decreases the average energy usage from approximately 62\% to 14\% compared to the \textit{Full Ensemble}, highlighting the potential for significant energy savings. In the case of the \textit{Dynamic Energy-Aware} strategy, we observed an average reduction from around 76\% to 57\%, showing the effectiveness of this approach in balancing performance with energy efficiency.

These findings confirm that model selection strategies, in their original form and particularly when augmented with an energy consumption metric, can combine the objectives of maintaining high accuracy while reducing the energy demands of AI systems in a production environment.

\vspace{-0.5em}
\section{Discussion}
This study presents several insights for strategic model selection for ensemble learning in live AI systems, particularly in performance optimization and computational efficiency. 

\subsection{(Green) AI}

For AI practitioners, the findings emphasize the importance and viability of balancing accuracy with computational cost. In the current context, where computational efficiency is both an economic and environmental concern, the study's insights suggest that achieving this balance—maintaining high levels of accuracy while being mindful of the computational resources consumed—is essential for sustainable AI development~\cite{schwartz2020green}. 

These insights provide practical industry benefits, particularly in strategizing AI developments. The potential for resource-efficient model selection without significantly compromising performance paves the way for greener AI solutions. Such solutions are particularly valuable in resource-intensive applications, contributing to efforts to reduce AI technologies' environmental footprint.

\subsection{Future of ensembles}

With the rising popularity of large language models (LLMs), one might wonder whether "old-school" ensembles are still the way to go. Research has shown that for most NLP tasks considered state-of-the-art fine-tuned models like TULRv6 generally outperform LLMs by a considerable margin, especially in languages other than English~\cite{ahuja2023mega}.  An LLM may not provide superior solutions for the specific task of legal information extraction.

In addition to accuracy, we prioritize efficiency as a crucial metric. A straightforward auto-regressive model like BERT$_{Large}$, which consists of approximately 340 million parameters~\cite{huggingface2020transformers}, requires four days of training on 16 TPU processors~\cite{huggingface2021bert}. In contrast, GPT-4 is rumored to have 1.7 trillion parameters and demands 90-100 days of training on 25,000 GPU processors~\cite{franca2023gpt4}. Although OpenAI has not officially disclosed the energy consumption of these models, it is reasonable to assume that their resource usage is significant. From the perspective of energy consumption, ensembles of fine-tuned models are preferable to large language models.

\vspace{-0.3em}
\subsection{Monitoring}
\label{subsection:monitoring}
Monitoring is critical to ensuring the sustainability of AI systems~\cite{henderson2020towards, wu2022sustainable}. Practitioners need to be able to monitor their applications to understand their energy consumption and environmental impact. Tools and interfaces that facilitate the development of AI systems should also incorporate features for monitoring energy usage. Monitoring tools would enable practitioners to create eco-friendly systems without significant effort, addressing the gap in making sustainable AI more accessible and practical. 

Moreover, by identifying the major energy consumers within live AI systems, practitioners can focus on reducing energy consumption in the most impactful areas, leading to more efficient and sustainable AI solutions. 
\vspace{-0.3em}
\subsection{Practical implications}

Balancing theoretical performance with practical considerations is vital when choosing the best model for a task. The \textit{Static} strategy is typically more straightforward to deploy and compatible with a broader range of infrastructures, making it a practical option for many setups. The \textit{Dynamic} strategy, while potentially more effective for specific tasks, might demand a more complex infrastructure setup. Therefore, the decision should consider the desired accuracy, efficiency, and practicality of integrating and maintaining the strategy into existing systems.

Companies should prioritize sustainable AI development as the impact of AI on the workforce continues to grow, accompanied by significant financial and environmental costs. Although the Green AI field is expanding, its connection to the industry remains insufficient~\cite{verdecchia2023systematic}. This research demonstrates that making AI systems in production more sustainable is feasible. Continued efforts in this direction can further convince companies of the practicality and benefits of adopting sustainable AI practices.

\subsection{Limitations}

While this study provides valuable insights, it also has limitations. The evaluation was conducted on two specific datasets within the legal domain, which may limit the generalizability of the findings to other types of documents or domains. Future research could explore the applicability of these model selection strategies across a broader range of datasets and domains, not only within information extraction but also in other areas beyond the scope of NLP.

Additionally, our approach to discounting model selection based on resource consumption relies on CPU seconds as a proxy for energy consumption. Future studies could incorporate more direct metrics of energy consumption, such as power usage in kWh, to assess the environmental impact and reduction in the carbon footprint of different model selection strategies. This limitation aligns with the continued need for accurate and easy-to-use monitoring tools for AI systems~\cite{henderson2020towards, wu2022sustainable} highlighted in Section \ref{subsection:monitoring}. 

Lastly, our study does not account for the energy costs associated with training or fine-tuning models for specific domains within the ensemble. Future research should investigate whether the accuracy improvements from domain-specific training, see blue models in Figure \ref{fig:models}, justify these increased energy expenditures. This analysis could provide deeper insights into the efficiency and effectiveness of employing specialized models within ensemble systems. 

\vspace{-1em}
\section{Related work}
Green AI, the intersection of energy efficiency and model accuracy in Artificial Intelligence (AI), has sparked a growing interest amongst researchers~\cite{strubell2019energy, schwartz2020green}. Our work focuses on reducing resource consumption in ensemble learning using model selection strategies in a live production environment. To our knowledge, there is no other work combining all elements. Below, we highlight the most significant contributions in model selection strategies and Green AI.

Zhou et al.~\cite{zhou2002ensembling} pivoted model selection in ensemble learning. Their work introduced GASEN, an approach that begins by allocating initial weights to neural networks and then employs a genetic algorithm to refine these weights. The optimized weights are instrumental in selecting the most effective subset of the ensemble. GASEN proves that incorporating the entire ensemble may not always be the optimal strategy, demonstrating the potential of selecting a subset of models. 

Li et al.~\cite{li2023towards} present a novel approach, IRENE, to ensemble learning that focuses on balancing performance and computational cost for inference of ensemble learning. Using a learnable selector, base models, and implementing early halting in a sequential model setup, IRENE reduces inference costs by up to 56\% while maintaining comparable performance to complete ensembles. IRENE presents a highly effective strategy for ensemble learning in contexts where sequential processing is feasible. However, the specific use case addressed in this paper necessitates a parallel approach and, as such, does not accommodate the sequential processing model that IRENE requires. 

David et al.~\cite{david2021adaptive} propose an ensemble learning approach based on the consensus of multiple models for class prediction. It operates under the assumption that once multiple models predict a class, it is likely the correct one. This approach significantly reduces computational costs by about 50\% while maintaining accuracy. Despite its effectiveness in classification tasks, this approach is not directly applicable to our work, as our ensemble is geared towards information extraction rather than classification, requiring a different methodological framework.  

Kotary et al.~\cite{kotary2023differentiable} introduce a framework that combines machine learning and combinatorial optimization for differentiable model selection in ensemble learning. Their method of storing data for predicting optimal subsets in a neural network contrasts with our approach, which involves a more straightforward collection and selection of models within an ensemble for task-specific optimization. Additionally, as with the method from David et al. highlighted above, the success achieved on classification tasks cannot immediately be transferred to information extraction.

Cordeira et al.\cite{cordeiro2023post} present a new approach called Post-Selection Dynamic Ensemble Selection (PS-DES). PS-DES evaluates ensembles selected by different DES techniques using different metrics to determine the best ensemble for each query instance. Their work introduces static and dynamic model selection based on evaluating a preliminary set of results. The relevance of these selection methods to our study stems from their design goal of integration into AI pipelines, which aligns with our focus on applying these strategies to a real-life use case. Consequently, we adopt these methods to reduce the number of models during inference in ensembles. 

Within this landscape of optimizing ensemble learning, the work of Gowda et al.~\cite{gowda2023watt} introduces a critical consideration for assessing model efficiency. They propose a GreenQuotientIndex (GQI) metric that penalizes high electricity consumption while considering accuracy. The authors conduct a comprehensive study on the electricity consumption of different deep learning models, highlighting the often overlooked trade-off between accuracy and energy efficiency. Our work takes this concept further by demonstrating the practical application of the GQI in real-time production environments that rely on model ensembles.  \newline

\vspace{-1.5em}
\section{Conclusion}


In response to the rising energy demands of AI-powered software (AIware), particularly those employing ensemble learning~\cite{li2023towards}, our empirical study investigates model selection strategies to optimize both accuracy and energy efficiency. 

Our research introduces and evaluates two model selection strategies, \textit{Static} and \textit{Dynamic}, aimed at optimizing the performance of ensemble learning systems while minimizing their energy usage.

By evaluating the \textit{Static} and \textit{Dynamic} strategies across the CUAD and Lease Agreement datasets, we have highlighted the adaptability and potential of these approaches to meet diverse needs. Our results reveal that the \textit{Static} strategy improves the F1 score beyond the baseline, reducing average energy usage from 100\% from the full ensemble to 62\%. 
The \textit{Dynamic} strategy further enhances F1 scores, while using on average 76\% compared to 100\% of the full ensemble. 

Additionally, we propose an approach that discounts accuracy with resource consumption, the \textit{Energy-Aware} approach, showing potential for further reducing energy usage without significantly impacting accuracy. This method further decreased the average energy usage of the \textit{Static} strategy from approximately 62\% to about 14\%, and for the optimal \textit{Dynamic} strategy, from around 76\% to roughly 57\%.

Our findings, especially the successful application of the \textit{Energy-Aware} approach, align with the principles of Green AI~\cite{schwartz2020green, strubell2019energy}, advocating for sustainable AI practices that maintain high-performance standards. 

This field study on \DocQMiner, an AI system actively used in the industry, highlights our research's real-world applicability and significance in advancing sustainable, efficient AI technologies for live production environments.

These insights provide a valuable perspective for the industry on developing AI in a resource-conscious yet effective manner. They highlight the feasibility of using model selection strategies to balance accuracy and computational efficiency, demonstrating the crucial need for adopting strategies that account for accuracy and environmental impact for ensuring sustainable development as AI progresses.

\newpage
\bibliographystyle{ACM-Reference-Format}
\bibliography{paper.bib}

\end{document}